\documentclass[lettersize,journal]{IEEEtran}
\usepackage{amsmath,amsfonts}
\usepackage{algorithmic}
\usepackage{algorithm}
\usepackage{array}
\usepackage[caption=false,font=normalsize,labelfont=sf,textfont=sf]{subfig}
\usepackage{textcomp}
\usepackage{stfloats}
\usepackage{url}
\usepackage{verbatim}
\usepackage{graphicx}
\usepackage{cite}
\usepackage{pifont}

\usepackage{graphicx}
\usepackage{float} 
\usepackage{multirow}
\usepackage{makecell}
\usepackage{graphicx}
\usepackage{caption}
\usepackage{url}
\usepackage{amsmath}
\usepackage{wrapfig}
\usepackage{array}
\usepackage{booktabs}
\usepackage{cite}
\usepackage{color}
\usepackage{xcolor}
\definecolor{citecolor}{HTML}{0071bc}
\usepackage[colorlinks, linkcolor=red,  anchorcolor=blue, citecolor=citecolor]{hyperref}

\usepackage[utf8]{inputenc} 
\usepackage[T1]{fontenc}    
\usepackage{hyperref}       
\usepackage{url}            
\usepackage{booktabs}       
\usepackage{amsfonts}       
\usepackage{nicefrac}       
\usepackage{microtype}      
\usepackage{xcolor}         
\usepackage[utf8]{inputenc}

\begin{document}

\title{Prompt-based Learning for Unpaired Image Captioning}


\author{\makecell{Peipei Zhu, Xiao Wang, \textit{Member, IEEE}, Lin Zhu, Zhenglong Sun, \textit{Senior Member, IEEE} \\ Wei-Shi Zheng, Yaowei Wang, \textit{Member, IEEE}, and Changwen Chen, \textit{Fellow, IEEE}}
\thanks{
Peipei Zhu is with the School of Science and Engineering, Chinese University of Hong Kong, Shenzhen 518172, China, and also with the Peng Cheng
Laboratory, Shenzhen 518066, China (e-mail: peipeizhu@link.cuhk.edu.cn).

Xiao Wang is with the School of Computer Science and Technology, Anhui University, Hefei 230601, China. (email: wangxiaocvpr@foxmail.com) 

Lin Zhu is with the School of Computer Science, Beijing Institute of Technology, Beijing 100081, China. (e-mail: linzhu@pku.edu.cn)

Zhenglong Sun is with the School of Science and Engineering, Chinese University of Hong Kong, Shenzhen 518172, China (e-mail: sunzhenglong@cuhk.edu.cn).

Wei-Shi Zheng is the School of Computer Science and Engineering, Sun Yat-Sen University, Guangzhou 510275, China, and also with with the Peng Cheng Laboratory, Shenzhen 518066, China (e-mail: wszheng@ieee.org).

Yaowei Wang is with the Peng Cheng Laboratory, Shenzhen 518066, China (e-mail: wangyw@pcl.ac.cn).

Changwen Chen is with the Department of Computing, The Hong Kong Polytechnic University, Hong Kong 999077, China (e-mail: changwen.chen@polyu.edu.hk).

Corresponding author: Yaowei Wang, Changwen Chen
}}




\maketitle

\begin{abstract}
Unpaired Image Captioning (UIC) has been developed to learn image descriptions from unaligned vision-language sample pairs. Existing works usually tackle this task using adversarial learning and visual concept reward based on reinforcement learning. However, these existing works were only able to learn limited cross-domain information in vision and language domains, which restrains the captioning performance of UIC.
Inspired by the success of Vision-Language Pre-Trained Models (VL-PTMs) in this research,
we attempt to infer the cross-domain cue information about a given image from the large VL-PTMs for the UIC task. This research is also motivated by recent successes of prompt learning in many downstream multi-modal tasks, including image-text retrieval and vision question answering. In this work, a semantic prompt is introduced and aggregated with visual features for more accurate caption prediction under the adversarial learning framework. In addition, a metric prompt is designed to select high-quality pseudo image-caption samples obtained from the basic captioning model and refine the model in an iterative manner. 
Extensive experiments on the COCO and Flickr30K datasets validate the promising captioning ability of the proposed model. We expect that the proposed prompt-based UIC model will stimulate a new line of research for the VL-PTMs based captioning.
\end{abstract}

\begin{IEEEkeywords}
Prompt-based Learning, Unpaired Image Captioning, Semantic Prompt, Metric Prompt
\end{IEEEkeywords}

\section{Introduction}

The goal of image captioning is to automatically describe visual images with natural languages. This is a cross-modality task that transfers information from the image domain to the language domain~\cite{wu2020fine, zhang2020integrating, yang2020captionnet}. With the release of large-scale captioning datasets \cite{vinyals2015show, vinyals2016show} and the advances in deep learning, the performance of image captioning has been continuously improved. It has been widely used in many applications, such as human-robot interaction \cite{das2017visual, hong2021vln}, visual aid for the blind \cite{wu2017automatic, gurari2018vizwiz, gurari2020captioning}, and automatic driving \cite{kim2018textual, wang2018look, omeiza2021explanations}. The mainstream image captioning models have followed the encoder-decoder paradigm~\cite{ben2021unpaired, del2020ratt}, which encodes the image into feature representation first and then decodes it into a sentence in a word-by-word fashion. Although the performance is good, such supervised learning based captioning models rely on massively labeled vision-language pairs \cite{yang2018multitask, zhang2018high, li2019know}, which is time- and energy-consuming. Also, the models trained on limited samples may have poor generalization ability.

\begin{figure*}[!bht]
	\centering
	\small
	\includegraphics[width=0.9\textwidth, height=3cm]{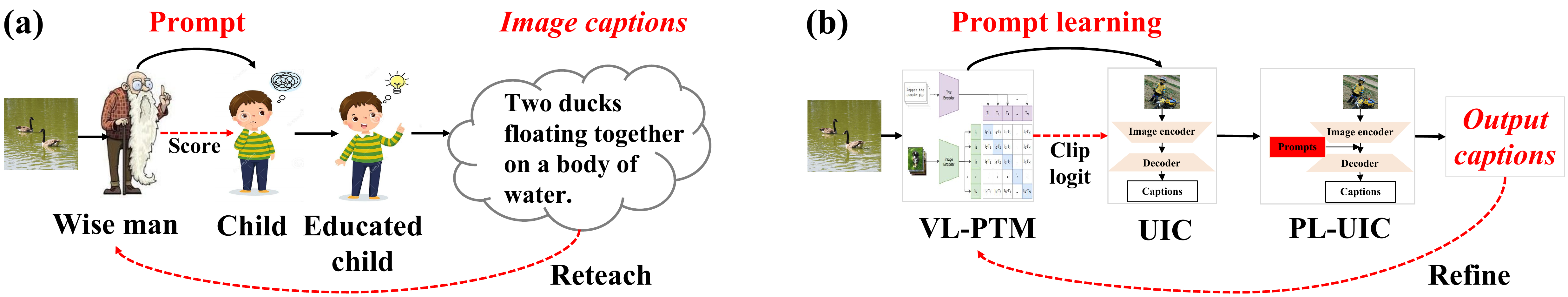}
	\caption{The prompt-based learning for unpaired image captioning. (a) An ignorant child learns knowledge from a wise man to describe an image. (b) The PL-UIC is developed to utilize prompts, learned from a vison-language pre-trained model (VL-PTM), to generate captions for images. These prompts of each image contain abundant contextual information of the matched images and texts, which is the information the previous UIC model does not have but is indispensable. The red dotted lines represent the process of caption scoring and reteaching in (a), which corresponds to the process of caption filtering and UIC model refining in (b).}
	\label{fig:first_image}
\end{figure*}


Considering the limitations of the fully-supervised image captioning paradigm, captioning using unpaired vision-language samples draws more and more attention as this approach does not require carefully labeled image-text training pairs. Usually, these models are developed based on \emph{adversarial learning}~\cite{chen2017show, donahue2016adversarial} and the \emph{visual concept reward} based on reinforcement learning \cite{feng2019unsupervised, laina2019towards}. As an early attempt, adversarial learning can only be utilized to guide the optimization of UIC parameters from the perspective of the overall structure of the sentence, while the correlations between the vision domain and the language domain have not been sufficiently explored. The concept reward based UIC models simply restrain their captions to contain the detected visual concepts (such as ``dog'' and ``tree''), therefore, their performance heavily depends on object detectors and very limited cross-domain knowledge is concerned. How to exploit more vision-language knowledge without paired image-text samples for UIC is still a challenging research problem to be resolved.
Recently, the pre-trained giant models~\cite{radford2021learning} have demonstrated their abundant prior knowledge by their superior performance in multiple domains and tasks, including natural language processing, computer vision, and multi-modal. These models carry an extremely large number of parameters and are pre-trained on the super large-scale corpus. For example, the CLIP~\cite{radford2021learning} is pre-trained with 400 million image-text pairs using cosine similarity maximization. Its superior performance on zero/few-shot learning demonstrates that it carries a lot of visual-language prior knowledge. Many other computer vision tasks have proved that the CLIP features further improve their performance significantly~\cite{wang2022clipnerf, zhang2022pointclip, narasimhan2021clip}. On the other hand, prompt learning~\cite{liu2021pre} is proposed to better leverage pre-trained models to improve overall performance on downstream tasks, such as PPT~\cite{gu2021ppt}, CoOp~\cite{zhou2022learning}, and VPT~\cite{jia2022visual}. 
These works inspire us to \emph{design new mechanisms for UIC by extracting prior vision-language knowledge from pre-trained big models.}

\begin{figure} 
	\centering
	\includegraphics[width=0.48\textwidth, height=2.3cm]{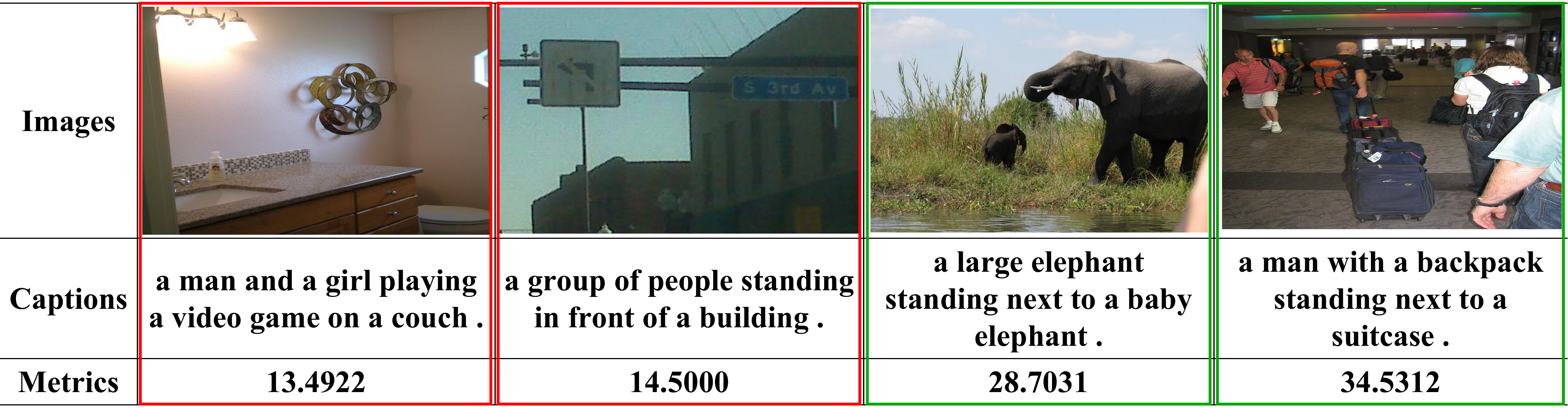}
	\caption{The metric prompt of image-caption pairs generated by the prompt-based UIC model. The higher value of the metric value, the higher quality of the image-caption pairs.}
	\label{fig:low_quality_captions}
\end{figure}

In this paper, a novel Prompt-based Learning scheme is proposed for UIC, termed PL-UIC, which can extract prior knowledge from the large-scale VL-PTMs. The key insight of this idea is similar to coaching a child to describe an image with the help of a wise man, as illustrated in Fig.~\ref{fig:first_image}. The child may describe the content of the given image more accurately if the wise man could give some important prompts. Therefore, two kinds of prompts are designed, \textit{i.e.}, the \emph{semantic prompt} and \emph{metric prompt}, to imitate such a learning paradigm. More specifically, the visual images are taken as input to the semantic prompt extraction module, consisting of the pre-trained VL-PTMs (CLIP~\cite{radford2021learning} is used in our experiments) and a feed-forward layer. The predicted prompt vector will be fed into the CLIP model to adjust its context and then align the image and prompt accurately. Then, the semantic prompt is injected into the adversarial learning based UIC framework for more intelligent and robust caption generation. 
The metric prompt is designed to transform the aforementioned unsupervised captioning optimization into a semi-supervised manner. As the pseudo captions can be obtained using the basic captioning model, and then high-quality samples can be filtered based on the metric prompt to polish the captioning model in an iterative way. As elaborated in Fig.~\ref{fig:low_quality_captions}, the metric value of an image and a caption obtained from the CLIP model can serve as the metric prompt. This semantic prompt-based learning and metric prompt guided high-quality sample filtering are integrated to form a strong caption generator~\emph{without using annotated aligned image-text pairs.}


To sum up, the contributions of this paper can be summarized as the following three aspects: 

$\bullet$ We have developed a novel Prompt-based Learning scheme for Unpaired Image Captioning, termed PL-UIC, which can make full use of VL-PTMs for high-performance captioning. To the best of our knowledge, it is the first work to infer the cue information (\textit{i.e.}, the prompt) about a given image that exists in the large VL-PTMs for the UIC task.

$\bullet$ Two types of simple yet effective prompt schemes have been designed for the UIC task, \textit{i.e.}, the semantic prompt and the metric prompt. The semantic prompt has been devised to extract vision-aware prior knowledge via the textual format and taken as input to guide the caption generation. The metric prompt guided pseudo label filter has been designed to help improving the selection of highly-matched image-caption pairs, which enabled us to enhance the proposed UIC model in a semi-supervised way.

$\bullet$ Extensive experiments have been carried out on the widely used COCO and Flickr30k datasets to demonstrate that the proposed prompt-based learning can efficiently boost the performance in caption generation. The design principle proposed in this research can also be applied to other applications that demand prior knowledge.

\section{Related Work}
In this section, we review the related works on supervised image captioning, unpaired image captioning models, and prompt learning.

\textbf{Image Captioning.~}
Classical image captioning implements the encoder-decoder architecture, which first encodes images into features and decodes these image features into sentences \cite{vinyals2015show, wu2019recall, xu2019multi} later. The goal of these models is to maximize the probability of generating the correct captions, relying on tremendous image-caption pairs \cite{hossain2019comprehensive, guo2019show}. To solve the problem of the tight dependence on the costly image-caption pairs, some researchers proposed to use fewer and fewer pairs to complete the task, including novel object captioning \cite{yao2017incorporating, hendricks2016deep} and semi-supervised image captioning \cite{chen2016semi, kim2019image}. 
Despite the promising captioning reform that has been completed, the costly paired image-caption datasets are indispensable in the training process. Distinct from all these works, we attempt to complete UIC without requiring any image-caption pair. 


\textbf{Unpaired Image Captioning.~}
Distinct from the aforementioned supervised image captioning, UIC is to generate descriptions for images without requiring any image-caption pairs. Feng \textit{et al.} \cite{feng2019unsupervised} tackled UIC via adversarial learning and the alignments between images and visual concepts. Although the UIC is achieved, the captioning performance has a big gap between UIC and supervised image captioning due to the weak vision-language correlations. Thus, some researchers put effort into enhancing the weak cross-domain correlations in the task \cite{laina2019towards, guo2021recurrent, ben2021unpaired}. For example, Laina \textit{et al.} \cite{laina2019towards} proposed to narrow the domain gap between images and languages by a shared embedding space of images and visual concepts. Also, several works focused on adopting scene graph modeling in UIC to align more textual information with images, including relationships and attributes \cite{gu2019unpaired, liu2019exploring, cao2020interactions, zhu2022unpaired}. The following methods achieved better captioning performance since much more vision-language alignment is explored in UIC. 
Despite the enhanced captioning performance, there is still much room for improvement due to the neglect of the majority of vision-language correlations. Different from all these works, we attempt to utilize the prompt-based learning in UIC, which is aided by the pre-trained CLIP model \cite{radford2021learning} with abundant vision-language prior knowledge.

\textbf{Prompt-based Learning.~}
Prompt-based learning methods are proposed in natural language processing (NLP), which aim to reduce or obviate the requirement for large supervised datasets in the downstream tasks \cite{liu2021pre}. When learning a language model, task-specific prompt functions are designed to model the probability of the text prompt itself, and this probability is then adopted to tackle the task \cite{wang2021learning, hu2021knowledgeable, rubin2021learning}.
During the training process, these prompt functions are utilized to instruct pre-trained models to perform corresponding tasks conditionally \cite{gu2021ppt, ding2021openprompt}. The developments of prompt-based methods make zero-shot and few-shot learning in NLP tasks come true \cite{brown2020language, schick2020exploiting}. Inspired by these developments,  researchers tried to extend it into vision tasks, such as image classification \cite{zhou2021learning}, visual question answering \cite{yao2021cpt, jin2022good}, image captioning \cite{jin2022good}, etc. 
In this work, to incorporate the vision-language prior knowledge into the UIC task, we propose a PL-UIC model which is inspired by the prompt-based learning. Distinct from all existing works, we take a two-step procedure: learning the semantic prompt of each image from a large pre-trained VL-PTMs and applying these prompts as additional guidance to generate captions firstly, and then one metric prompt is utilized to design a high-quality pseudo label filter to further enhance the captioning performance.


\section{Methodology}
In this section, we will first introduce an overview of the proposed PL-UIC. Then, the unpaired image captioning is reviewed briefly. After that, the motivation of semantic prompt and metric prompt is discussed carefully. Later, the designed semantic prompt for UIC is introduced in detail. Finally, we discuss the metric prompt guided pseudo label filter for UIC model polish.

\subsection{Overview} 
\begin{figure*}[t]
\centering  
\small
\includegraphics[width=0.94\textwidth,height = 6.4cm]{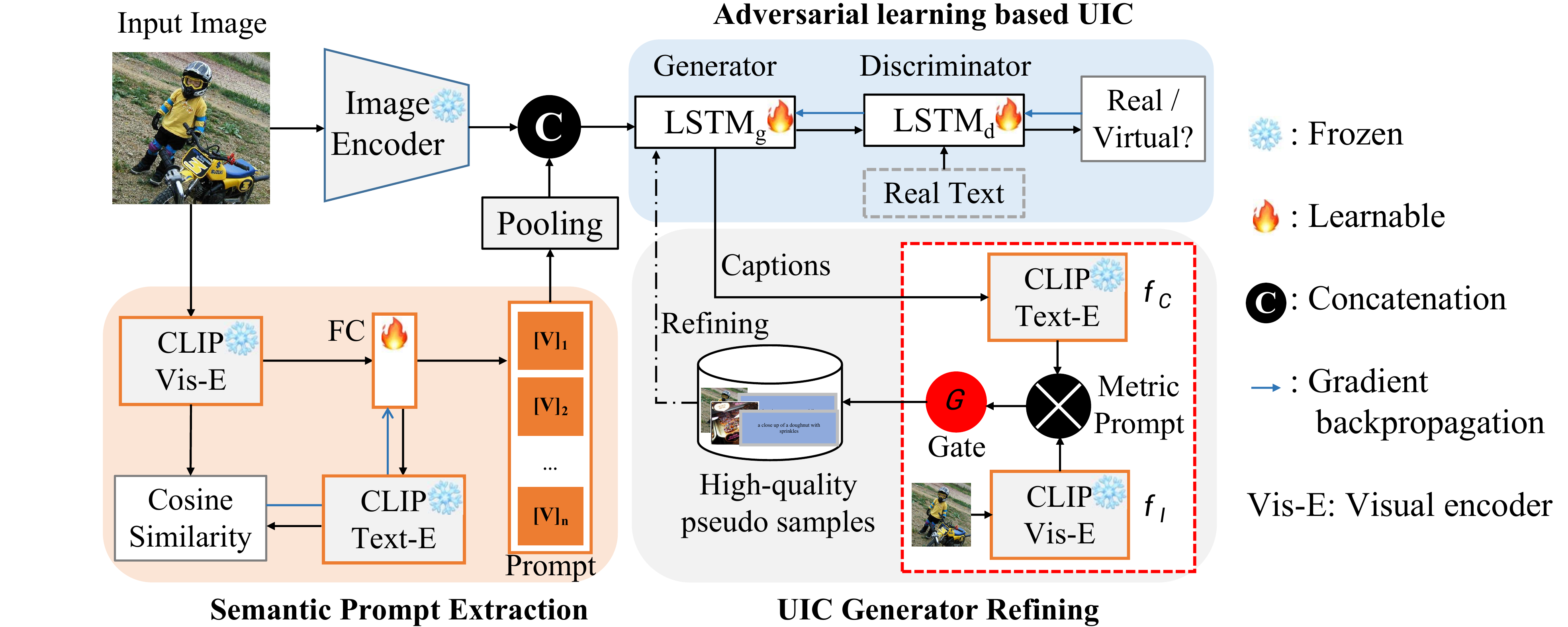}
\caption{An overview of the proposed PL-UIC framework, which contains three modules, \textit{i.e.}, the semantic prompt extraction, the adversarial learning based UIC, and the UIC generator refining module. Firstly, the designed semantic prompt extraction module, a feed-forward layer with the frozen CLIP model, is trained to generate the semantic prompt for each image. Then, these prompts are implemented as guidance in the adversarial learning based UIC. Finally, the proposed pseudo label filtering is utilized to re-train the UIC generator iteratively, which selects high-quality image-caption pairs as the pseudo samples through the measurement of the metric prompt.}
\label{fig:Overall_Framework}
\end{figure*}

The target of PL-UIC is to train an unpaired image captioning model guided by prompts. As shown in Fig.~\ref{fig:Overall_Framework}, the overall framework consists of three parts, \textit{i.e.}, the semantic prompt extraction module, the adversarial learning based UIC model, and the metric prompt based generator refining module. Specifically, the semantic prompt extraction module contains the CLIP model and a feed-forward layer to provide the semantic prompt for each image in the image datasets. Then, the semantic prompt guided UIC model is trained to generate captions by adopting unpaired image-text samples. Later, the caption generator refining module is trained using pseudo labels. These pseudo labels are selected by the designed metric prompt, which can measure the correlations of image-text pairs. We combine the designed two prompt-related modules into the common adversarial learning UIC framework, which achieves promising results on two widely used captioning datasets.

\subsection{Preliminary: Unpaired Image Captioning}

\begin{figure}[t]
\centering  
\small
\includegraphics[width=0.48\textwidth,height = 1.75cm]{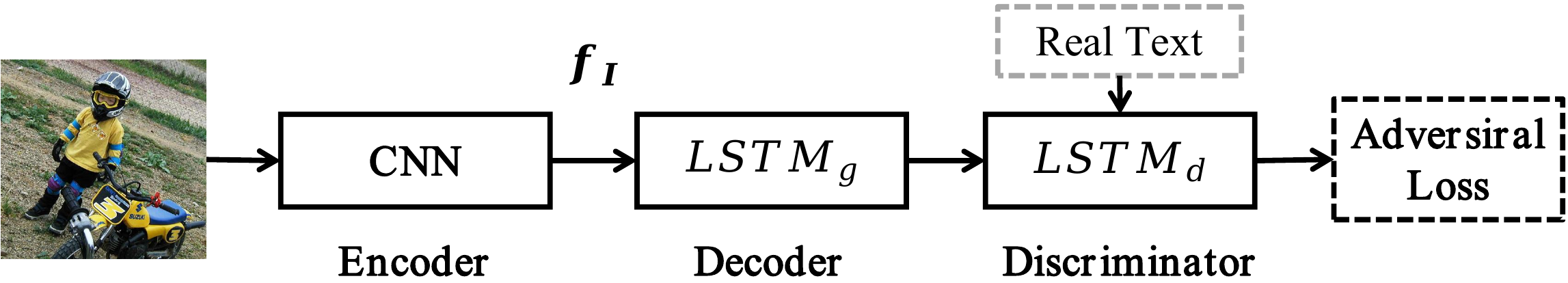}
\caption{The adversarial learning for UIC.}
\label{fig:encoder_decoder}
\end{figure}

Unpaired image captioning is to describe images without using any aligned vision-text pairs. The image dataset with $N_i$ images is represented by $\mathcal{I}=\{\textit{\textbf{I}}_1,\textit{ \textbf{I}}_2,..., \textit{\textbf{I}}_{N_{i}}\}$, the unpaired language dataset is represented as $\mathcal{S}=\{\textit{\textbf{S}}_1, \textit{\textbf{S}}_2,..., \textit{\textbf{S}}_{N_{s}}\}$ with $N_s$ sentences, and the captions generated by the UIC model are denoted as $\mathcal{C}=\{\textit{\textbf{C}}_1, \textit{\textbf{C}}_2,..., \textit{\textbf{C}}_{N_{c}}\}$, where $N_c$ means the number of generated captions. For simplicity, we utilize $\textit{\textbf{I}}$, $\textit{\textbf{C}}$, and $\textit{\textbf{S}}$ to represent an image, a virtual generated caption, and a real sentence, respectively. 

The common pipeline of UIC is illustrated in Fig. \ref{fig:encoder_decoder}, where a CNN-LSTM network is adopted for the encoder-decoder framework. Firstly, a CNN encoder is utilized to extract the image features $\textit{\textbf{f}}_\textbf{I}$. 
Then, an LSTM decoder network $LSTM_g$  is adopted to transform the image features into texts   $\textit{\textbf{C}} = \{\textit{\textbf{c}}_1, \textit{\textbf{c}}_2, ..., \textit{\textbf{c}}_{n_c}\}$, where $n_c$ is the number of words in one caption. And this procedure can be written as: 
\begin{equation} \textit{\textbf{x}}_{0} =  FC(\textit{\textbf{f}}_\textit{\textbf{I}}),
\end{equation}
  \begin{equation} 
  \textit{\textbf{x}}_{t} = \textit{\textbf{W}}_e \textit{\textbf{c}}_{t}, t \in \{1, ..., n-1 \},
  \label{xt}
  \end{equation}
  \begin{equation}
     [\textit{\textbf{o}}_{t+1}, \textit{\textbf{h}}_{t+1} ]=  LSTM_{g}(\textit{\textbf{x}}_{t},\textit{ \textbf{h}}_{t}), t \in \{0,\ ...,\  n -1\} 
     \label{lstmg}
  \end{equation} 
  \begin{equation}
      \textit{\textbf{c}}_t \sim \textit{\textbf{o}}_{t}, t \in \{1, ..., n \}
      \label{co}
  \end{equation}
where $\textit{\textbf{x}}_t$, $\textit{\textbf{c}}_t$, $\textit{\textbf{o}}_t$, and $\textit{\textbf{h}}_{t}$  are the input of the $LSTM_g$ decoder layer,  a one-hot vector representation of the outputted word, the probability of every word in the dictionary, and the hidden state of the LSTM at the $t$-th time step, separately. $\textit{ \textbf{x}}_{0}$ is the initial input of the decoder, and one feed-forward layer $FC$ is utilized to adjust the image features. $\textit{ \textbf{h}}_{0}$ is a zero vector for the initial hidden states of the decoder. $\textit{\textbf{W}}_e$ is for word embedding. Besides, an LSTM discriminator $LSTM_d$  is adopted to differentiate a real sentence $\textit{\textbf{S}}$ and a virtual sentence $\textit{\textbf{C}}$, which is utilized to make the generated captions as real sentence as possible. Formally,
\begin{equation}
[\textit{q}_{t}, \textit{\textbf{h}}_{t} ]= LSTM_{d}(\textit{\textbf{x}}_{t}, \textit{\textbf{h}}_{t-1}), t \in \{1,\ ...,\  n\},
\label{lstmd}
\end{equation}
where $\textit{q}_t$ represents the probability that the outputted sequential words $\{\textit{\textbf{c}}_1,...,\textit{ \textbf{c}}_t\}$ of one caption $\textit{\textbf{C}}$ or $\{\textit{\textbf{s}}_1, ..., \textit{\textbf{s}}_{t}\} $ from a real sentence $\textit{\textbf{S}}$ are regarded as a real sentence $\textit{\textbf{S}}$.

Although the adversarial learning based UIC models work well in some scenarios, however, their overall performance is still limited due to the less cross-domain knowledge. We think this situation can be alleviated by introducing vision-language prior knowledge. 

\subsection{Motivation of Semantic Prompt and Metric Prompt}

The motivation of the semantic prompt and metric prompt is the vision-language prior knowledge fully reflected in the feature representation of the VL-PTMs. For example, the CLIP model used in this paper performs well in zero/few-shot learning since its abundant language-aware visual features. The CLIP model includes one visual encoder and one sentence encoder. The visual encoder is used to extract the deep image features, which can be implemented with ResNet-101 \cite{he2016deep} or ViT \cite{dosovitskiy2020image}. The text encoder is used to learn the sentence features using a Transformer network \cite{parmar2018image}. Based on a large amount of image-text dataset (400 million) crawled from the Internet, the CLIP is optimized to align features of these two domains using contrastive learning \cite{chen2020simple}. Specifically, its target is to maximize the cosine similarity between the matched image-caption pairs and minimize the unmatched ones. Therefore, the cross-domain vision-language prior knowledge can be mastered by the CLIP. 

\begin{figure*} [t!]
	\centering
	\small
	\subfloat[]{
		\includegraphics[height=70mm,width=80mm]{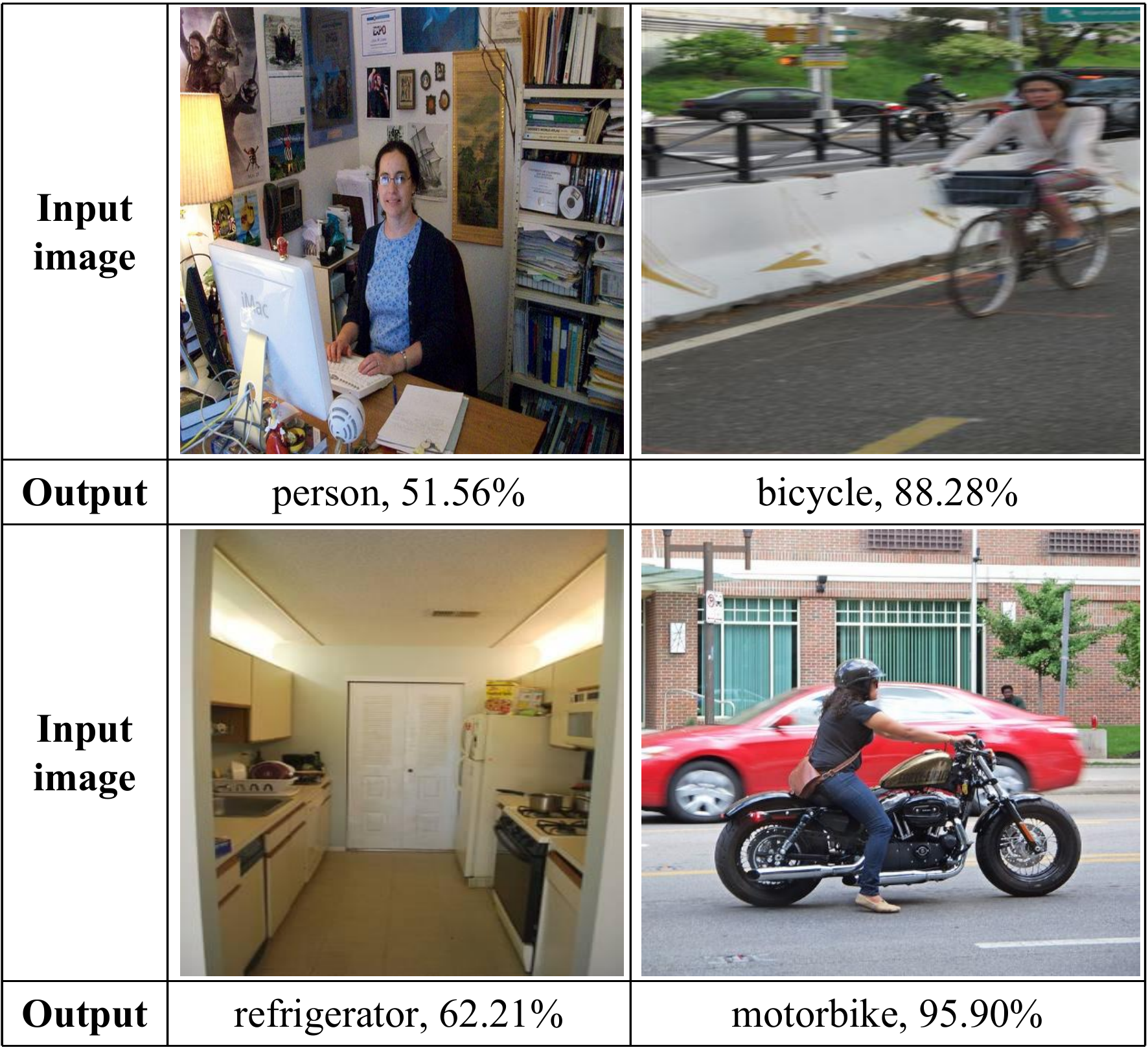}}
	\subfloat[]{
		\includegraphics[height=70mm,width=80mm]{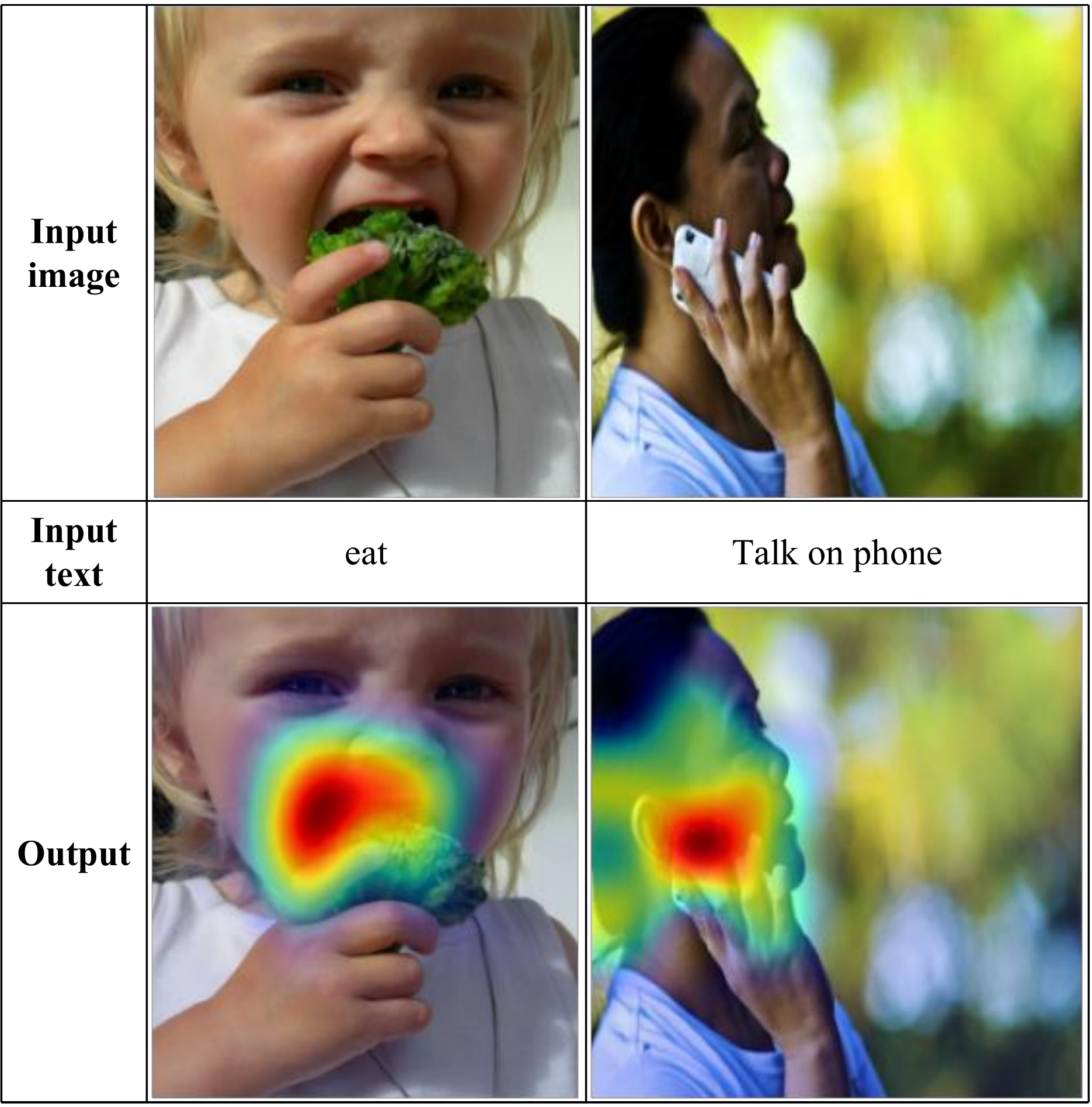}}
	\caption{The illustration of the vision-language prior knowledge in the CLIP model. (a) Examples of the zero-shot image classification, and (b) The highlighted heatmaps for an input caption without training.}
	\label{fig:prior_knowledge} 
\end{figure*}

To intuitively exhibit the prior knowledge of the CLIP model, two types of examples are visualized in Fig. \ref{fig:prior_knowledge}. Fig. \ref{fig:prior_knowledge} (a) is the zero-shot image classification, which reflects the prior knowledge from the vision domain to the language domain. Fig. \ref{fig:prior_knowledge} (b) is about highlighting the most related regions in an image for the given texts without training, which illustrates the prior knowledge from the language domain to the vision domain.



The main target of this paper is to extract the learned knowledge from CLIP for the guidance of unpaired image captioning from three aspects, \textit{i.e.}, the visual encoder, the semantic prompt, and the metric prompt. The visual encoder of the CLIP model is adopted to extract the image features for UIC directly, where abundant language-aware image features are provided to guide the caption generation. The details of the two prompts are introduced in the following sections.

\subsection{Semantic Prompt for Unpaired Image Captioning}

For unpaired image captioning, the performance of existing methods is restrained by the limited cross-domain knowledge. To alleviate the issue, a semantic prompt extraction module is designed to explore more cross-domain prior knowledge from the VL-PTMs. The prompt extracted from the module will be taken as guidance for the caption generation.

\noindent\textbf{Semantic Prompt Extraction} is used to draw vision-aware textual knowledge from pre-trained vision-language models, \textit{i.e.}, the semantic prompt for UIC task. The semantic prompt is represented by $\hat{\textit{\textbf{P}}_i}\in \{\hat{\textit{\textbf{P}}}_1, \hat{\textit{\textbf{P}}}_2,..., \hat{\textit{\textbf{P}}}_{N_{p}}\}$ for each image $\textbf{\textit{I}}_i$ in the image datasets. For simplicity, we utilize $\hat{\textit{\textbf{P}}}$ to represent a semantic prompt. Different from the prior knowledge obtained from the fixed visual encoder, the semantic prompt is obtained by training with the frozen CLIP model via given visual images. In detail, the features from the image encoder of the CLIP model are taken as the input, as shown in Fig. \ref{fig:Overall_Framework}. For practical implementations, we adopt one feed-forward layer to achieve this goal. Then, the output of the feed-forward layer $FC$ will be fed into the text encoder for subsequent processing. Formally, 
\begin{equation}
    \textit{\textit{\textbf{p}}} =  FC(\textit{\textbf{f}}_\textit{\textbf{I}}),
\end{equation}
\begin{equation}
\textit{\textbf{f}}_\textit{\textbf{p}} = TE(\textit{\textbf{p}}),
\end{equation}
where $\textit{\textbf{p}} \in \mathbb{R}^{y}$ means the output of the $FC$ layer. $y$ denotes the dimension of the semantic prompt, and it will be reshaped into the length of the prompt and the dimension of each prompt, such as $8 \times 512$. $TE$ means the CLIP text encoder and $\textit{\textbf{f}}_\textit{\textbf{p}}$ means the encoded prompt features. Note that the parameters of the image and text encoder of the CLIP model are frozen. The objective of this module is to maximize the cosine similarity of the image features $\textit{\textbf{f}}_\textit{\textbf{I}}$ and the matched prompt features $\textit{\textbf{f}}_\textit{\textbf{p}}$, and minimize the cosine similarity of unmatched pairs. The cosine similarity $\hat{S}$ can be written as:
\begin{equation}
  \hat{S} = \frac{\sum_{i} \textit{\textbf{f}}_\textit{\textbf{I}} \cdot \textit{\textbf{f}}_\textit{\textbf{p}}}{||\textit{\textbf{f}}_\textit{\textbf{I}}||\times ||\textit{\textbf{f}}_\textit{\textbf{p}}||}, 
\end{equation}
where $\sum_i$ means the summation of all elements in $\textit{\textbf{f}}_\textit{\textbf{I}} \cdot \textit{\textbf{f}}_\textit{\textbf{p}}$. 

To obtain the semantic prompt $\hat{\textit{\textbf{P}}}$ for each image, the ``SOS'', ``CLS'', and ``EOS'' embeddings are concatenated at the start and end of the FC layer, which represents the start and end of the prompts, respectively. Moreover, the positional pointers of the CLIP model are added to the former embeddings, and the results $\hat{\textit{\textbf{P}}}$ are taken as the learned prompt for each image. Formally,
\begin{equation}
\hat{\textit{\textbf{P}}} = [\textit{\textbf{E}}_{SOS}, \textit{\textbf{p}}, \textit{\textbf{E}}_{CLS},\textit{ \textbf{E}}_{EOS}] + \textit{\textbf{E}}_{pos},
\end{equation}
where $[\cdot]$ means the concatenation operation, and $\textit{\textbf{E}}$ represents various types of embeddings in CLIP. The $\textit{\textbf{E}}_{pos}$ indicates the positional embeddings in the CLIP model. And we utilize ${[V]_1, [V]_2, ..., [V]_n}$ to represent a semantic prompt $\hat{\textit{\textbf{P}}}$, where $[V]_1$ denotes the features for a textual word. Note that only the image dataset of UIC is utilized for training this module, and the architecture of the trainable layer in the semantic prompt extraction module can be replaced by other networks for better performance, which we left in the future work.

\noindent\textbf{Caption Generation} To integrate the semantic prompt $\hat{\textit{\textbf{P}}}$ into the UIC task, we take the prompts of each image as part of the input of the UIC network. Firstly, a pooling block $Pool$ is adopted to reduce the computation related to the prompts. Then, the prompts $\textit{\textbf{f}}_{\hat{\textit{\textbf{P}}}}$ and image features $\textit{\textbf{f}}_\textit{\textbf{I}}$ are concatenated together and are taken as the input features of the caption generator $LSTM_g$ at the first time step $\textit{\textbf{x}}_{0}$ to guide the caption generation. Formally,
\begin{equation}
    \textit{\textbf{f}}_{\hat{\textit{\textbf{P}}}} = Pool(\hat{\textit{\textbf{P}}})
\end{equation}
\begin{equation}
    \textit{\textbf{x}}_{0} = [\textit{\textbf{f}}_\textit{\textbf{I}}, \textit{\textbf{f}}_{\hat{\textit{\textbf{P}}}}].
\end{equation}
Then, the sentence decoder $LSTM_g$ is utilized to generate texts based on $\textit{\textbf{x}}_{0}$ and the process takes the same principle as in the common UIC model, as shown in Equation~\ref{xt}, \ref{lstmg}, and \ref{co}. After that, a virtual text will be inputted into the discriminator $LSTM_d$ to differentiate whether it is real or not, shown in Equation \ref{lstmd}. Through the training of sentence decoder and discriminator, we will obtain relatively accurate captions for each image $\{\textit{\textbf{I}}, \textit{\textbf{C}}\}$ through the trained semantic prompt-based UIC model.

\subsection{Metric Prompt for UIC Generator Refining}

\begin{figure}[!tb]
\centering  
\small
\includegraphics[width=0.46\textwidth,height = 8.1cm]{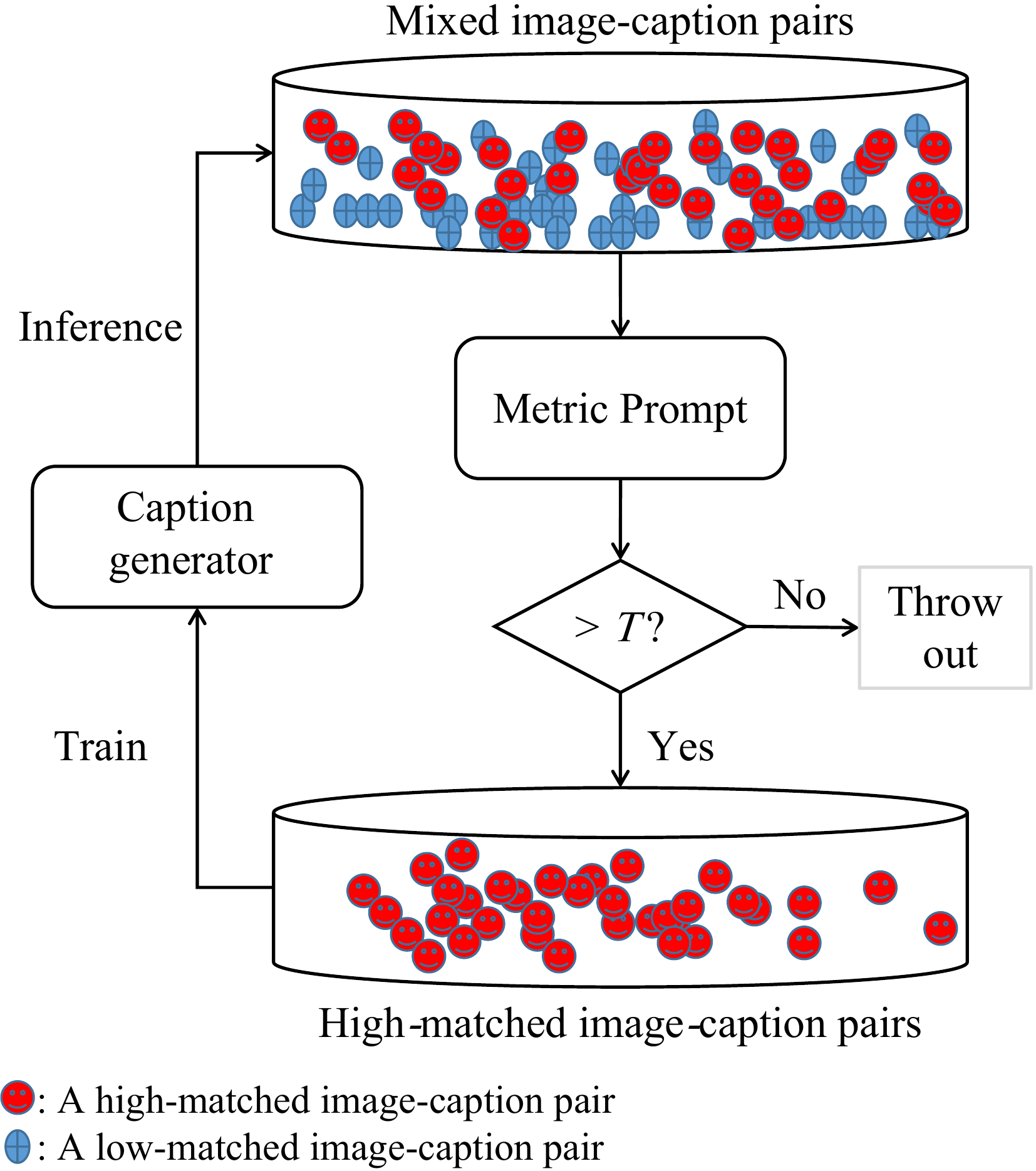}
\caption{The process of the caption generator refining. The generator first outputs mixed image-caption pairs, including low-matched ones and high-matched ones. Then, the metric prompt of one image-caption pair is computed and compared with a threshold $T$. Suppose the metric prompt is higher than the $T$, the image-caption pair will be regarded as a high-matched pair and be selected to train the caption generator again.}
\label{fig:metric_prompt}
\end{figure}

After the training of semantic prompt-based UIC is done, we can get better results than the prompt-free UIC model. Due to the limitation of unsupervised learning of captioning generator, the unpaired image-text samples are still hard to optimize, which bring us sub-optimal performance only. Therefore, we are inspired to transform the unpaired image-text samples into paired ones using generated captions for the training images. Then, these data can be utilized as pseudo labels to train a fully-supervised caption generator. 

Considering that the UIC model is still weak in the initial stage, indiscriminate leveraging these data may harm our model. In this section, a pseudo label filtering scheme is proposed to filter out the low-quality image-caption pairs and leave the high-quality ones, which enables us to polish our model in a more stable way. Based on the proposed scheme, the model refining and label filtering processes can be executed iteratively. The whole process is illustrated in Fig. \ref{fig:metric_prompt}. In the filtering process, the metric prompt is designed to select the high-matched image-caption pairs. Unlike the prior knowledge in the semantic prompt, the metric prompt contains different cross-domain knowledge that indicates whether an image and a caption are matched.


\noindent\textbf{Metric Prompt.} The metric prompt $L$ is adopted as a criterion to filter out the low-quality image-caption pairs, which represents the cosine similarity between an image and a caption. Specifically, we input the visual image and its caption into the pre-trained CLIP model. Therefore, we can get the prompt via: 
\begin{equation}
   L =  \textit{\textbf{w}} \times (\frac{\sum_i\textit{\textbf{f}}_\textit{\textbf{I}} \cdot \textit{\textbf{f}}_\textit{\textbf{C}}}{||\textit{\textbf{f}}_\textit{\textbf{I}}||\times ||\textit{\textbf{f}}_\textit{\textbf{C}}||} ),
\end{equation}
where $\textit{\textbf{w}}$ and $\textit{\textbf{f}}_\textit{\textbf{C}}$ are the weights and the caption features, respectively. 

As illustrated in Fig. \ref{fig:low_quality_captions}, the high-quality image-caption pair has a high metric value, and the low-quality image-caption pair has a low metric value. Suppose we have a metric gate $G$ with threshold $T$, if the metric is lower than the threshold $T$, the raw image caption $\{\textit{\textbf{I}}, \textit{\textbf{C}}\}$ will be filtered out. Otherwise, the raw image caption pair will be preserved to train the semi-supervised caption generator.

\noindent\textbf{Iterative Sample Filtering and Caption Generator Refining.} In the experiments, we conduct the captioning generator refining and sample filtering in an iterative way. As the captioning generator will output higher quality captions when a new iterative loop is completed (\textit{i.e.}, the sample selection and generator re-training), therefore, it can in turn contribute to training a better semi-supervised caption generator. The filtering and re-training procedure can be executed multiple times to further enhance the captioning ability of the generator. We provide the experimental analysis of the influence of the iteration number in Section \ref{iterationResults}.

\subsection{Training and Inference}
\label{Sec:T_I}

\noindent\textbf{Training Phase.} 
In this work, we apply a multi-stage optimization scheme for the network since the proposed framework consists of multiple modules which are required to be optimized one by one. 


\textbf{\emph{Stage-I: }} The semantic prompt extraction module, \textit{i.e.}, a single FC layer embedded on the CLIP model, is firstly trained on an image dataset. As a result, the semantic prompt of each image will be obtained, which represents the contextual features of images and texts, and will be utilized as a part of the input of the UIC model. 

\textbf{\emph{Stage-II: }} Aided by the aforementioned semantic prompts, the UIC model is trained by the image features and semantic prompts, which is optimized by adversarial learning and visual concept alignments. The trained model can be adopted to generate the captions of images.

\textbf{\emph{Stage-III: }} With the generated image-text pairs, the captioning model is re-trained to enhance the captioning performance. In this stage, metric prompt-based pseudo label filtering is implemented to select the high-quality image-text pairs as the pseudo labels. These labels can be utilized to train a better semi-supervised model. The filtering and re-training strategies loop multiple times to enhance the captioning performance.

\noindent\textbf{Inference Phase.} 
Given the testing images, we first feed them into the semantic prompt extraction module to obtain the semantic prompt. Then, these images and corresponding prompts are inputted into the trained semi-supervised model to obtain high-quality captions.

\section{Experiments}

\subsection{Datasets and Evaluation Metrics}

\noindent\textbf{Datasets.} 
Different from the standard image captioning task, which utilizes paired image-text samples, the UIC task adopts unpaired datasets. In our experiments, five datasets are involved, including two image datasets (\textit{i.e.}, the COCO~\cite{lin2014microsoft} and Flickr30K~\cite{lu2017knowing}) and three language datasets (\textit{i.e.}, the COCO captions, Shutterstock sentences~\cite{feng2019unsupervised}, and Conceptual sentences~\cite{sharma2018conceptual}).

A brief introduction to these datasets is given below. 1). COCO~\cite{lin2014microsoft} is one of the popular benchmarks of the image captioning tasks, which contains 123,287 images. The common Karpathy split~\cite{karpathy2015deep} is utilized in the experiments, with 113,287 training images, 5,000 validation images, and 5,000 testing images. 2). Flickr30K dataset consists of 31,783 images with 29,783 training images, 1,000 validation images, and 1,000 testing images, which is the common split~\cite{lu2017knowing} for the experiments on image captioning. 3). COCO caption dataset is annotated manually. Each image is annotated with 5 captions. For training, the COCO images and captions are utilized in an unpaired way. A word vocabulary is built by words presented no less than 4 times in captions of the training images. 4). Shutterstock sentence dataset is crawled from the Shutterstock website by~\cite{feng2019unsupervised}, which contains 2,282,444 different image captions. 5). Conceptual sentences~\cite{sharma2018conceptual} are automatically collected from webs, which include 3.3 million image-caption pairs. The captions are utilized as a sentence dataset.

\begin{table}
\caption{Comparisons of the proposed PL-UIC with related methods on the testing split of COCO dataset for UIC. SP: semantic prompt. * represents the re-trained version of the related works with the same powerful CLIP visual encoder. ``(finetuning SP)'' means that all parameters in the semantic extraction module are trainable. ``(3)'' represents three loops of metric prompt based model refining.}
\centering
    \small
    \begin{tabular}{c|ccccccc}
    \hline
    Method &B4 & M & R & C  &  S\\ \hline
      \makecell{Feat2sen  \cite{feng2019unsupervised}* }           &15.7&18.0&41.5&48.7&10.8\\
    \makecell{Recons-Align  \cite{feng2019unsupervised}* }         &22.3& 21.5 &	47.1& 71.9& 14.4\\
     \makecell{WS-UIC  \cite{zhu2022unpaired}* }         &22.6& 20.9 &	46.6& 69.2& 14.1\\ \hline
    \makecell[c]{PL-UIC (finetuning SP) (3)}& 24.3 & 22.3 & 48.7 &75.6 & 14.5\\
    \makecell[c]{PL-UIC (3)}& 24.9 & 22.5 &  49.3& 77.3&14.9 \\ 
\makecell[c]{PL-UIC}& \textbf{25.0} & \textbf{22.6} & \textbf{49.4}& \textbf{77.9} &\textbf{15.1} \\ \hline
    \end{tabular}
    \label{tab:comparsion_with_sota}
\end{table}

\noindent\textbf{Evaluation Metrics.}
Five evaluation metrics, including BLEU-4 (B4)~\cite{papineni2002bleu}, METEOR (M)~\cite{denkowski2014meteor}, ROUGE (R)~\cite{lin2004looking}, CIDEr (C)~\cite{vedantam2015cider} and SPICE (S)~\cite{anderson2016spice}, are adopted to evaluate the captioning performance, whose values are computed based on the ground-truth captions of test images. Among these metrics, BLEU-4~\cite{papineni2002bleu} was proposed for machine translation by computing the overlapping units of the machine translations and human translations in 2002; language-specific evaluation was brought to evaluate machine translation of any language in 2014 and the metric is called METER~\cite{denkowski2014meteor}; ROUGE~\cite{lin2004looking} was mainly designed for automatic abstracting by counting the overlapping units of computer-outputted summaries and human labeled summaries in 2004; in 2015, CIDEr~\cite{vedantam2015cider} was proposed for evaluating image captioning by capturing human judgment of consensus; and the semantic propositional content of human caption evaluation was considered in SPICE~\cite{anderson2016spice} for evaluating image captioning in 2016. The higher the metric values, the better the experimental performance of UIC.

\subsection{Implementation Details}
In UIC, we utilize the pretrained ViT-b16 CLIP model as the image encoder. The object concepts are extracted by Faster-RCNN~\cite{ren2016faster} related object detection model. The LSTM with 512 hidden states is utilized for the sentence decoder and discriminator. The learning rate of the caption generator is set as 0.001 at the training \textbf{Stage II} in Section~\ref{Sec:T_I}. At the training \textbf{Stage III}, the learning rate is set as 0.00001. The architecture of the semi-supervised image captioning model is the same as the generator of the UIC model. The whole PL-UIC model is trained on 2 V100 GPUs. 

For the semantic prompt extraction and pseudo label filtering, the ViT-b16 visual encoder~\cite{radford2021learning} is utilized in the CLIP model. For pseudo label filtering of COCO image datasets, the threshold of metric prompt is set as 30. For the Flickr30k image dataset, the threshold of the metric prompt is set as 26 due to the smaller scale of the image dataset.

\subsection{Comparison on Benchmarks}

\begin{table}
\caption{Comparisons with related methods for UIC using independent datasets. * represents the re-trained version of the related works with the same powerful CLIP visual encoder as the image encoder in UIC.}
\centering
    \small
    \begin{tabular}{cccccccc}
    \hline
    Method &B4 & M & R & C  &  S\\ \hline
    \multicolumn{6}{c}{ \textbf{COCO images + Shutterstock sentences}}\\
    \makecell[c]{Feat2sen \cite{feng2019unsupervised}*}& 5.5 & 12.7& 28.8 & 24.8 & 8.4 \\
    \makecell[c]{Recons-Align \cite{feng2019unsupervised}*}& 6.6&12.9 & 28.0 &31.3  &8.9 \\
    \makecell[c]{WS-UIC \cite{zhu2022unpaired}*}& 6.4&12.4 & 29.2 &26.7  &7.9 \\ \hline
\makecell[c]{PL-UIC}& \textbf{10.0} &\textbf{16.2}& \textbf{35.8}&\textbf{45.8}&\textbf{11.5} \\ \hline \hline
 \multicolumn{6}{c}{ \textbf{Flickr30k images + Conceptual sentences}} \\
    \makecell[c]{Feat2sen \cite{feng2019unsupervised}*} &7.8&\textbf{11.4}&29.7&9.5&\textbf{6.5}\\
    \makecell[c]{Recons-Align \cite{feng2019unsupervised}*} & 5.2& 10.1& 26.2 & \textbf{11.4} &5.7\\
     \makecell[c]{WS-UIC \cite{zhu2022unpaired}*} &6.2&10.3&32.4&7.6&4.7\\ \hline
\makecell[c]{PL-UIC}& \textbf{9.7} & 10.9&\textbf{32.7} & 8.8 & \textbf{6.5}\\ \hline
\end{tabular}
\label{tab:perf_compare_Flickr}
\end{table}

In this section, two kinds of experimental comparisons are carried out to validate the effectiveness of the proposed approach. The \textbf{first} one is to compare the PL-UIC with other approaches on the COCO dataset using unpaired image and caption samples. The \textbf{second} setting is to compare PL-UIC with algorithms adopting independent datasets, \textit{i.e.}, COCO images with Shuttershock sentences, and Flickr30k images with Conceptual captions. We compare the PL-UIC with the related methods Feat2sen \cite{ feng2019unsupervised}, Recons-Align of \cite{ feng2019unsupervised}, and WS-UIC \cite{zhu2022unpaired}. Feat2sen relies on pseudo labels, outputted by an object-to-sentence model \cite{feng2019unsupervised}, to train a fully-supervised caption generator, where only sparse knowledge of the text domain and the image domain is concerned. Recons-Align depends on adversarial learning and the visual concept reward, both of which cannot be implemented to explore the contextual vision-language information thoroughly. WS-UIC relies on one more unrecognized object loss to improve the alignment between the objects and images, which is still weak to explore plentiful of cross-domain knowledge. As a result, the experimental performance of these methods is severely constrained. Different from all these works, we adopt the semantic prompt and metric prompt of each image, containing full of vision-language prior knowledge, as additional guidance for caption generation.


\begin{table}
\caption{Ablation study to verify the effectiveness of each module in the proposed PL-UIC. W/o P: without using VL-PTMs. VP: vision encoder of VL-PTMs. SP: semantic prompt. MP: metric prompt. MP(1): metric prompt with 1 iteration. MP(3): metric prompt with 3 iterations.}
    \centering
    \small
    \setlength{\tabcolsep}{1mm}{
\begin{tabular}{cccccccccc}
    \hline 
    \multicolumn{5}{c}{Method} & \multicolumn{5}{c}{Evaluation Metric} \\ 
    W/o P & VP & SP &MP (1)& MP (3) & B4 & M & R & C  &  S\\ \hline
    \makecell[c]{\ding{51}} & & & & & 18.4&18.3&43.3&56.7& 11.5 \\ 
    \makecell[c]{\ding{51}} & & \ding{51}& & &19.0 &18.9 &43.9 & 58.6&  11.6\\ 
\makecell[c]{}& \ding{51} & & & & 20.7& 20.1 &	45.9& 64.4& 12.7 \\ 
\makecell[c]{}& \ding{51}& \ding{51} & & & 21.6	&20.6&	46.1&	67.2&	13.4 \\ 
\makecell[c]{}& \ding{51}&  &\ding{51} & &24.0 & 22.1& 48.0	&74.8	&14.0		 \\ 
\makecell[c]{}&\ding{51} &\ding{51} &\ding{51} & & 24.2& 22.1  &48.7 &75.1& 14.7 \\ 
\makecell[c]{}& \ding{51}&\ding{51} & &\ding{51} &\textbf{24.9}  & \textbf{22.5}&\textbf{49.3}&\textbf{77.3}&\textbf{14.9} \\
\hline
\end{tabular}}
\label{tab:ablation_study}
\end{table}
 
\begin{figure} [t!]
	\centering
	\small
	\subfloat[]{
		\includegraphics[scale=0.28]{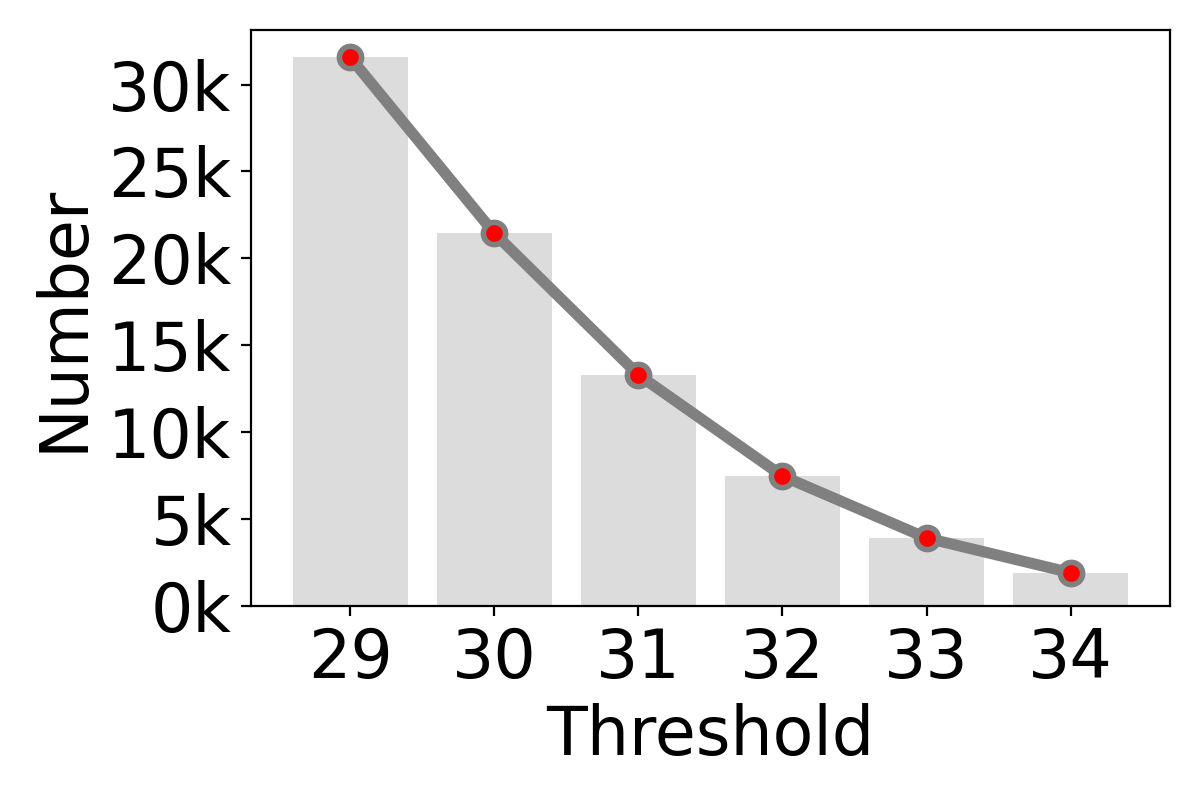}}
	\subfloat[]{
		\includegraphics[scale=0.28]{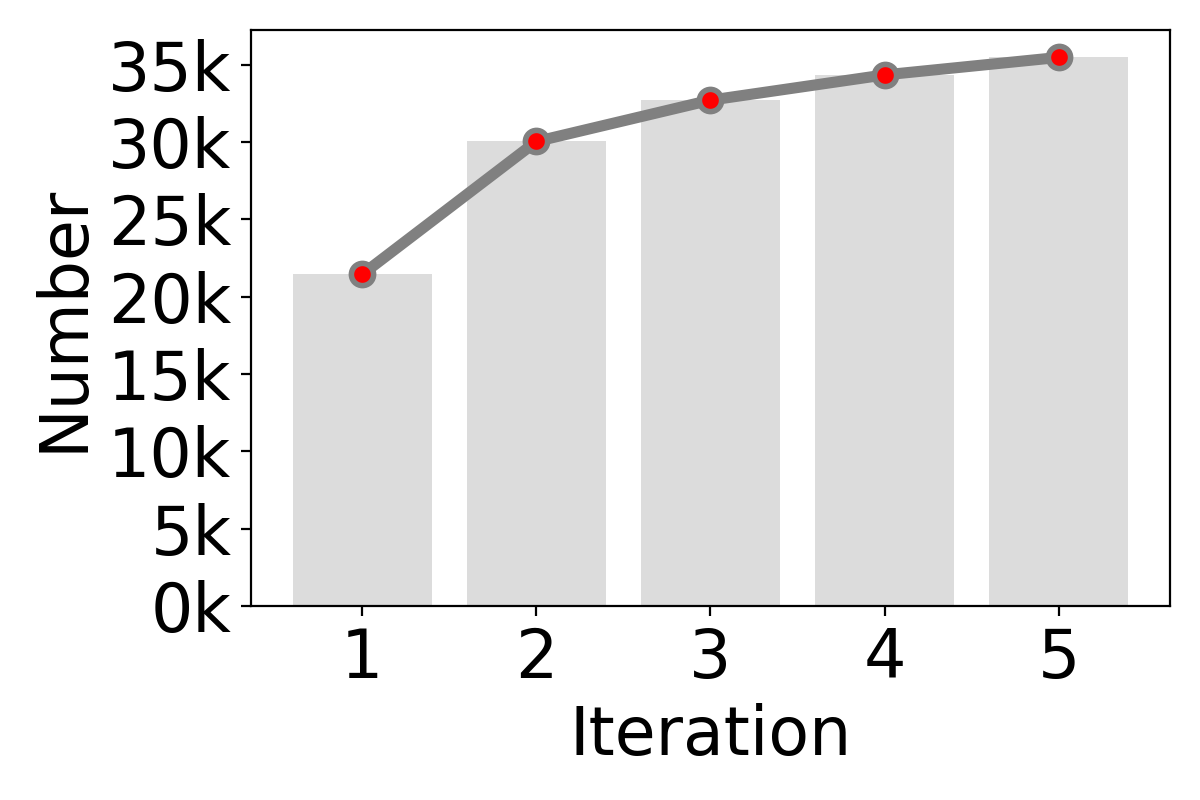}}
	\caption{The number of pseudo labels in (a) different thresholds of the metric gate and (b) different iterations of the metric prompt related model refining. The threshold of the metric gate is set as 30 in (b).}
	\label{fig:number_of_pseudo_labels} 
\end{figure}



For the first setting, we report the experimental results of ours and the related UIC models in Table \ref{tab:comparsion_with_sota}. * means the re-trained version of methods with the same powerful image encoder of PL-UIC. Besides the method Feat2sen, Recons-Align, and WS-UIC, we compare with the PL-UIC (finetuning SP) (3) and PL-UIC (3). The PL-UIC (finetuning SP) (3) means training PL-UIC with finetuned semantic prompts and 3 iterations of the generator refining, where the semantic prompt are extracted by training the whole model of the semantic prompt extraction instead of one feed-forward layer. PL-UIC (3) means that the experiment is carried out by 3 iterations of the caption generator refining. Generally speaking, the experimental results of the proposed PL-UIC achieve the best captioning performance over all five evaluation metrics. For example, the C value of PL-UIC is 6\% higher than the C value of Recons-Align \cite{ben2021unpaired}. The results basically demonstrate that the extracted prompts, with plentiful contextual vision-language information, can provide efficient guidance for captioning. And the iterative pseudo label filtering scheme can provide high-quality labels for generator refining. Compared the PL-UIC (finetuning SP) (3) with PL-UIC (3), the higher experimental results of PL-UIC (3) verify that the training of only one feed-forward layer of the designed semantic prompt extraction method is more effective than training all parameters of the whole module.

For the second setting, we compare the proposed approach with related methods by utilizing independent datasets. From Table \ref{tab:perf_compare_Flickr}, it is easy to find that our proposed PL-UIC outperforms all the compared methods by a significant margin in the experiments with COCO images and Shutterstock sentences. More in detail, the C value of PL-UIC is 45.8. Meanwhile, the value of Recons-Align is only 31.3, which is 14.5 lower than PL-UIC. For the experiments with Flickr30k images and Conceptual sentences, the overall captioning performance of PL-UIC exceeds all the compared methods, although it has weak performance on metrics M and C. These experimental results fully demonstrate that our PL-UIC model has the strongest captioning ability in the independent dataset settings since the vision-language prior knowledge learned from VL-PTMs. 

\begin{figure*}[!ht]
    \small
	\centering
	\includegraphics[width=1\textwidth, height=2.8cm]{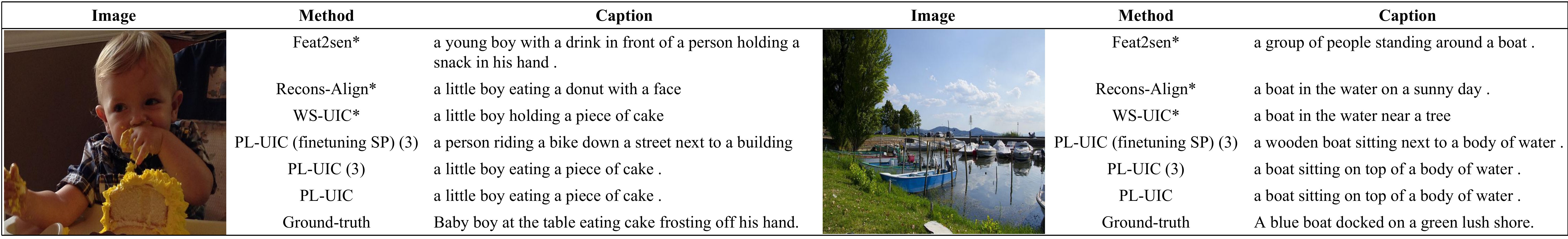}
	\caption{Examples of the captions generated by different methods on the COCO dataset with ground-truth captions.}
	\label{fig:captions_compare}
\end{figure*}

To compare the captioning ability of all these methods more intuitively, two images with the corresponding captions generated by these methods are elaborated in Fig. \ref{fig:captions_compare}. From the figure, we can observe that the captions outputted by the proposed PL-UIC are better than the other compared methods. Take the first image as an example, the Feat2sen* method outputs "a young boy with a drink in front of a person holding a snack in his hand.'', where only the concept "a young boy" is accurate. The method Recons-Align* and WS-UIC* generate incorrect concept ``donut'' and "holding", respectively.  PL-UIC (finetuning SP) (3) generates a totally wrong caption ``a person riding a bike down a street next to a building''. The captions outputted by the proposed method PL-UIC (3) and PL-UIC are semantically correct with "a little boy", "eating", and "a piece of cake", which contain almost all the important concepts as in the ground-truth captions. Visualizing these captions intuitively demonstrates the superiority of the proposed PL-UIC with plentiful cross-domain prior knowledge.

\subsection{Ablation Studies}

\subsubsection{Component Analysis} 
To facilitate readers to have a more thorough understanding of the proposed approach, extensive experiments for the component analysis have been conducted on the testing split of the COCO dataset. As shown in Table \ref{tab:ablation_study}, the following modules are implemented or not for the comparisons: ``W/o P'' means utilizing an image encoder from a relatively small off-the-shell model instead of the VL-PTMs; ``VP'' is adopting the visual encoder of CLIP as the image encoder of UIC; ``SP'' means the semantic prompt; ``MP'' denotes the metric prompt; ``(1)'' represents 1 iteration of the generator refining; and ``(3)'' means 3 iterations of the generator refining.



As shown in Table \ref{tab:ablation_study}, we can find that the model ``W/o P'' only achieves 18.4, 18.3, 43.3, 56.7, and 11.5 on the B4, M, R, C, and S metric, respectively. The experimental results of the method ``W/o P'' are obviously lower than the method ``VP'', which demonstrates the effectiveness of the language-aware visual prior knowledge from the CLIP model. 
When introducing the semantic prompt to the ``W/o P'', the overall results can be improved to 19.0, 18.9, 43.9, 58.6, and 11.6, respectively. Also, when the other two methods ``VP'' and ``VP'' with ``MP(1)'' are combined with ``SP'', the experimental performances are also better than the methods without ``SP''. These results validate that our designed semantic prompt based on VL-PTMs is beneficial for the UIC task due to the abundant vision-aware language prior knowledge.
When we integrate the metric prompt-related module, our results can be further improved. Specifically, the model with ``MP(1)'' is better than the ``non-MP(1)'' version. 
For instance, the B4, M, R, C, and S can be improved up by +3.3, +2.0, +2.1, +10.4, and +1.3, respectively, when one iteration of the metric prompt-related module is used with ``VP''. Obviously, more iterations can bring better results. These results fully validate the effectiveness of our proposed modules for the UIC task due to the prior metric knowledge of the VL-PTMs. We hope our model can bring new insights to exploring contextual vision-language information for the VL-PTMs or prompt learning-based UIC.

\subsubsection{Effect of Different Image Encoders}
To investigate the effects of the language-aware prior knowledge from different vision backbones, we conduct the UIC experiments with three image encoders from different vision backbones, \textit{i.e.}, ResNet-101, ViT-b32, and ViT-b16, illustrated in Table \ref{tab:vision_backbones}. Moreover, the experiments with Inception-V4, not related to the VL-PTMs, are carried out for comparison. In the table, we can observe that all these three VL-PTMs related experiments achieve more promising captioning performance than the non-VL-PTMs experiments. For example, the VL-PTMs related ResNet-101 experiments perform 20.2, 20.1, 45.0, 63.3, and 12.8 over metric B4, M, R, C, and S, respectively, while Inception-V4 related experiments only achieve 19.0, 18.9, 43.9, 58.6, and 11.6 over the same metrics, respectively. These experiments show that the language-aware vision prior knowledge in the image encoders is effective and efficient for the UIC task, no matter what type of vision backbone it is.

\subsubsection{Effect of Semantic Prompt}

\textbf{Prompt Length.} To explore the influence of the length of the semantic prompt, three lengths are utilized to carry out the experiments, \textit{i.e.}, 4, 8, and 16. As exhibited in Table \ref{tab:prompt_len}, these three experiments have comparable experimental performance over evaluation metrics B4, M, R, and S. For example, these three experiments obtain 21.4, 21.6, and 21.6 over metric B4, respectively. As for metric C, the experiment of the semantic prompt with length 8 has a slight advantage over the other two experiments. In detail, experiments with prompt length 8 achieve 67.2 on metric C, while the experiments with prompt length 4 and 16 obtain only 66.8 and 65.5 on the same metric, respectively. We can conclude that the prompt length has a relatively bigger influence on the evaluation metric C than other metrics. And we choose the prompt length 8 in other types of semantic prompt-related experiments.

\noindent\textbf{Prompt from Different Vision Backbones.} To verify the effectiveness of the semantic prompt learned from different vision backbones, three different backbones, including ResNet-101, ViT-b32, and ViT-b16, are utilized in the experiments to extract the semantic prompt, respectively. For comparison, we carry out the experiments with the same image encoder but without the semantic prompt (W/o Prompt). As elaborated in Table \ref{tab:prompt_of_different_vision_backbone}, all these semantic prompt-related experiments elaborate promising captioning performance over five evaluation metrics than the non-prompt experiments, demonstrating the generality of vision backbones for the semantic prompt extraction. For instance, the experimental results with the semantic prompt learned from the ResNet-101 backbone of the CLIP model achieve 21.1, 20.4, 45.7, 66.8, and 13.5 on metrics B4, M, R, C, and S, respectively, while the experiments without semantic prompt only perform 20.7, 20.1,	45.9, 64.4, and 12.7 over the same evaluation metrics, respectively.

\begin{table}
\caption{The effect of the image encoder from different vision backbones.}
\centering
\small
\begin{tabular}{c|cccccc}
 \hline
  \specialrule{0em}{-0.2pt}{-0.2pt}
    Method & B4 & M & R & C  &  S\\ \hline
    \makecell{InceptionV4 }  &19.0&18.9&43.9&58.6&11.6 \\ \hline
    \makecell[c]{ResNet-101}& 20.2 &20.1&45.0&63.3&12.8 \\ 
    \makecell[c]{ViT-b32}& 19.8 &19.2&44.8&58.6&12.0\\ 
    \makecell[c]{ViT-b16}& 21.3	&20.4&	45.8&	66.5&	13.5\\ \hline
    \end{tabular}
\label{tab:vision_backbones}
\end{table}

\begin{table}
\caption{The effect of the semantic prompt from different vision backbones.}
\centering
\small
\begin{tabular}{c|cccccc}
 \hline
  \specialrule{0em}{-0.2pt}{-0.2pt}
    Method & B4 & M & R & C  &  S\\ \hline
   \makecell[c]{W/o Prompt}&20.7& 20.1 &	45.9& 64.4& 12.7 \\ \hline
    \makecell[c]{ResNet-101}& 21.1 &20.4&45.7&66.8&13.5 \\ 
    \makecell[c]{ViT-b32}& 21.1 &20.7 &46.0&66.4&13.6\\ 
    \makecell[c]{ViT-b16}& 21.3	&20.4&	45.8&	66.5&	13.5\\ \hline
    \end{tabular}
\label{tab:prompt_of_different_vision_backbone}
\end{table}

\begin{table}
 \caption{The effect of prompt length.}
\centering
\small
\begin{tabular}{cccccc}
\hline
 Length & B4 & M & R & C & S \\
\hline
\makecell{4} & 21.4 & \textbf{20.6} & 46.0& 66.8& \textbf{13.6}\\
\hline 
 8 & \textbf{21.6}  &\textbf{20.6} &\textbf{46.1}&\textbf{67.2}&13.4\\
\hline
 16 & \textbf{21.6}  &20.3 &45.9&65.5&13.0\\ \hline
 \specialrule{0em}{-0.2pt}{-0.2pt}
 \end{tabular}
\label{tab:prompt_len}
\end{table}

\subsubsection{Effect of Metric Prompt}

\begin{figure*}[ht]
    \centering
    \includegraphics[width=1\textwidth, height=2.6cm]{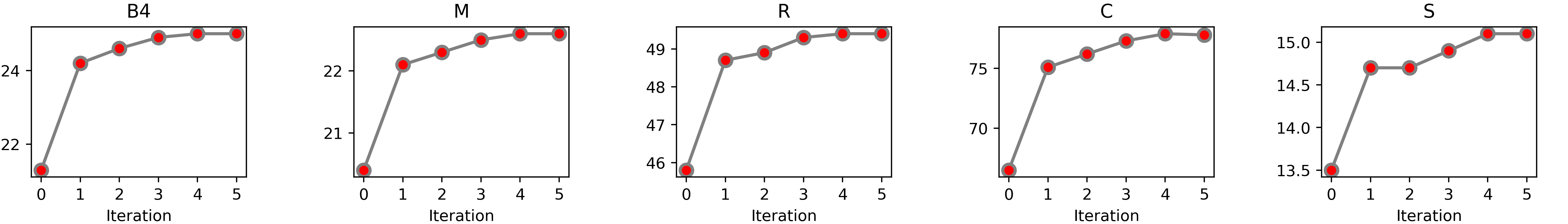}
    \caption{The iteration influence of the pseudo label filtering on different evaluation metrics, including the B4, M, R, C, and S.}
    \label{fig:iteration}
\end{figure*}

\textbf{Iterative Generator Re-training. }  \label{iterationResults} 
The multiple iterations of the generator re-training are indispensable for training a better generator since higher-quality labels will be collected in the next loop. From Fig. \ref{fig:number_of_pseudo_labels} (b), we can observe that the number of pseudo labels increases with the number of iterations. As more high-quality labels can be used in the supervised learning phase, the captioning performance is getting better and better, as shown in Fig. \ref{fig:iteration}. The values are higher and higher over all five evaluation metrics, including B4, M, R, C, and S. These two figures demonstrate the usefulness of the iteration scheme. 

\noindent\textbf{Threshold.}
The pseudo labels for UIC generator refining are generated by comparing the metric prompt of an image and a text with a predefined threshold.
To achieve better performance, we set different thresholds in this experiment, as shown in Table \ref{tab:threshold}. The higher the threshold, the less the number of pseudo labels, as illustrated in Fig. \ref{fig:number_of_pseudo_labels} (a). If the threshold equals 0, it means utilizing all generated image-caption pairs as the pseudo labels to refine the caption generator without filtering. Compared to the experiments with non-zero thresholds 29, 30, 31, and 32, the experimental results with 0 threshold are much worse. Since the threshold 33 and 34 are so high, the number of pseudo labels become very small with less dataset variety which limits the refining performance. Obviously, we can find that when the threshold is set as 30, the best captioning results can be obtained as 24.2, 22.1, 48.7, 75.1, and 14.7 for evaluation metrics B4, M, R, C, and S, respectively.  


\begin{table}
\caption{The effect of different thresholds of the metric prompt.}
\centering
\small
\begin{tabular}{c|cccccc}
 \hline
    Threshold & B4 & M & R & C  &  S\\ \hline
    \makecell[c]{0}& 22.7  &20.7&46.6&67.2&13.3 \\ \hline
    \makecell[c]{29}& \textbf{24.2}  &22.0&48.5&74.2&14.5 \\ \hline
    \makecell[c]{30}& \textbf{24.2} &\textbf{22.1} &\textbf{48.7}&\textbf{75.1}&\textbf{14.7}\\ \hline
    \makecell[c]{31}&23.8&22.0&48.6&74.4&14.6\\ \hline
    \makecell[c]{32}&23.6&21.8&48.3&72.8&14.2\\ \hline
    \makecell[c]{33}&22.0&20.8&47.0&67.6&13.3\\ \hline
    \makecell[c]{34}&20.6&20.0&45.9&61.7&12.4\\ \hline
    \end{tabular}
\label{tab:threshold}
\end{table}

%

\begin{figure*}[!ht]
	\centering
	\includegraphics[width=1\textwidth, height=10cm]{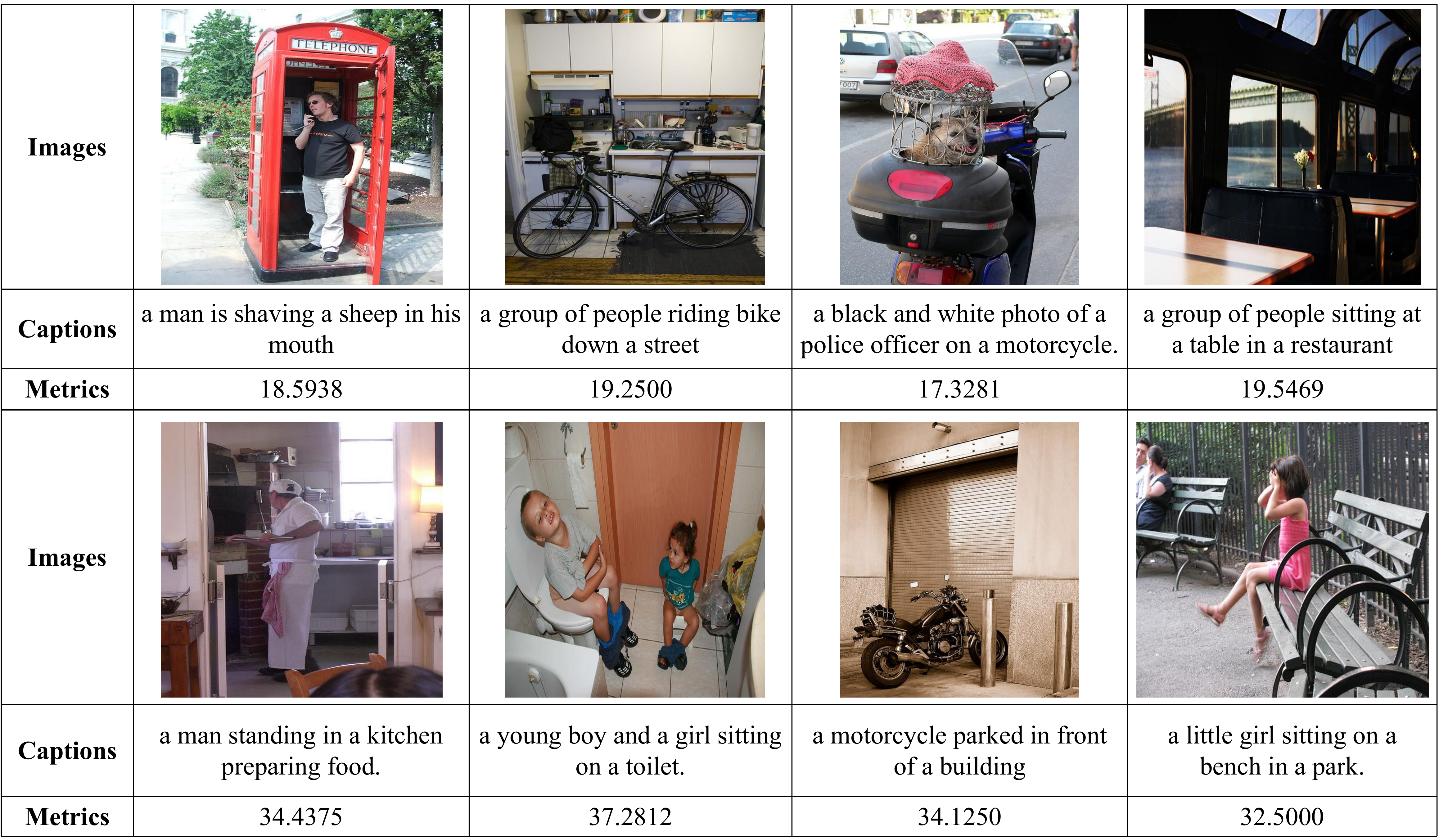}
	\caption{The metric value of image-caption pairs.}
	\label{fig:logits}
\end{figure*}

\begin{figure*}[!ht]
	\centering
	\includegraphics[width=1\textwidth, height=10cm]{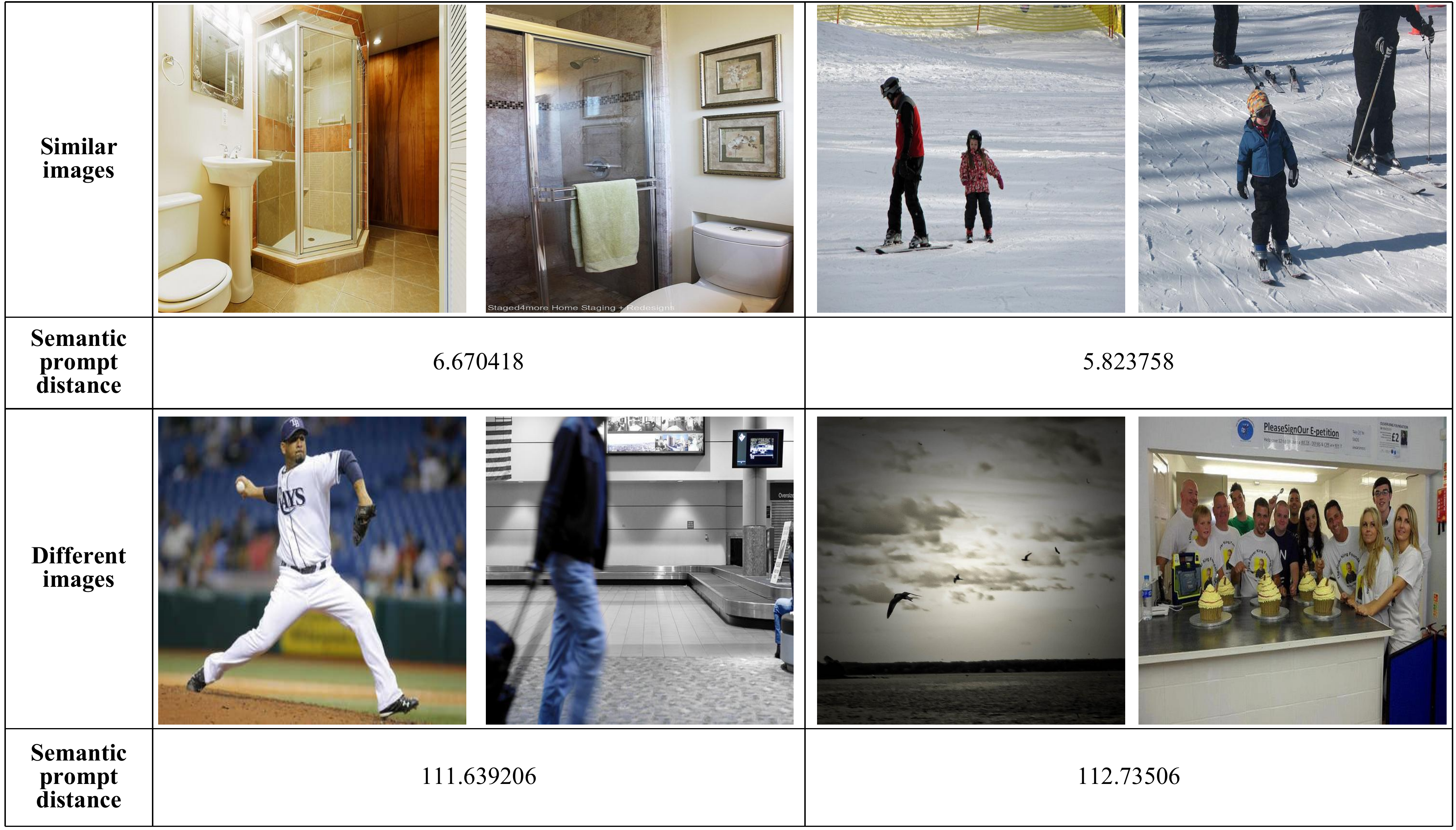}
	\caption{The distance of semantic prompts between similar images and different images, respectively.}
	\label{fig:prompt_dis}
\end{figure*}

\begin{figure*}[!ht]
    \small
	\centering
	\includegraphics[width=1\textwidth, height=8cm]{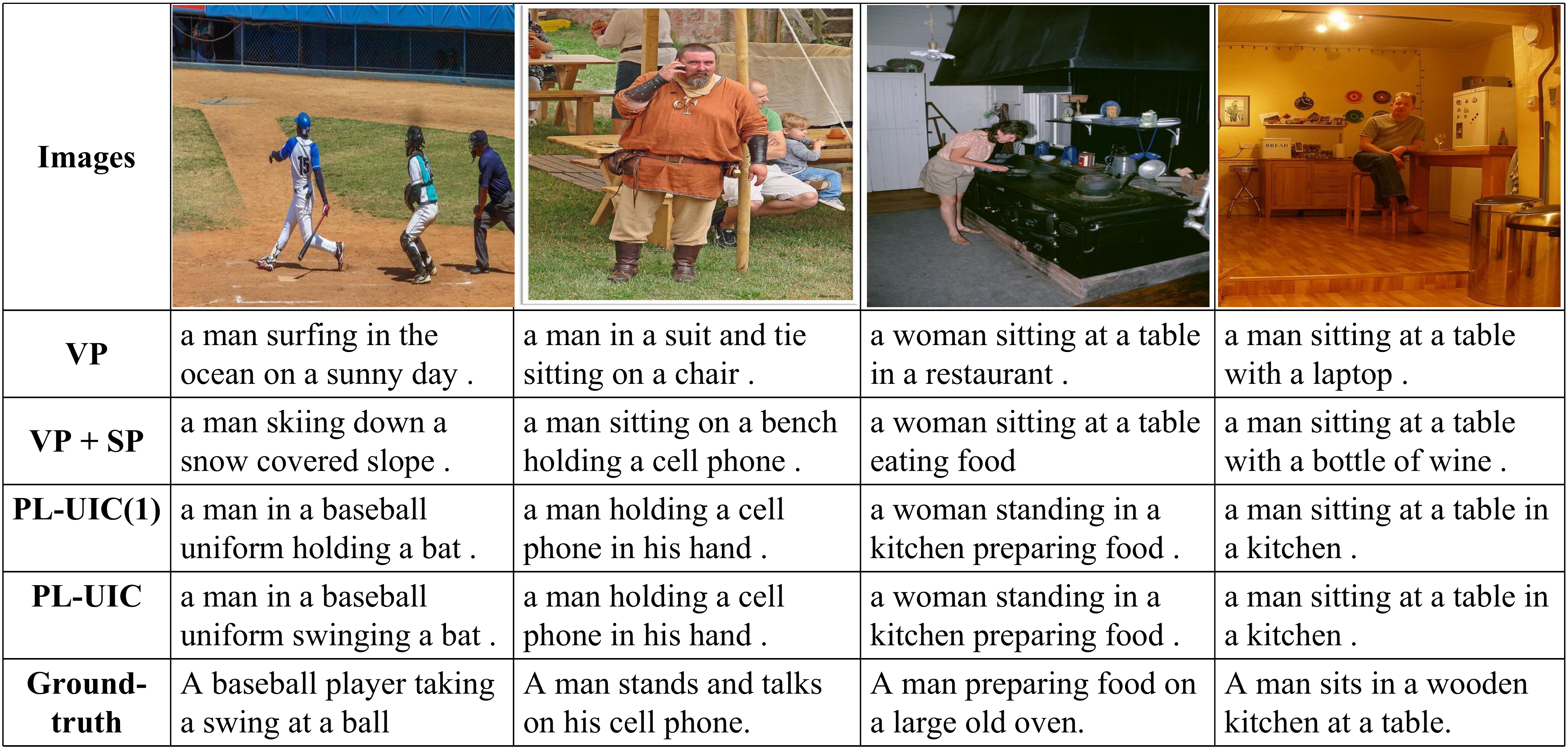}
	\caption{Examples of the generated captions on the COCO dataset with ground-truth captions.}
	\label{fig:captions}
\end{figure*}

\subsection{Qualitative Results}

\noindent\textbf{Visualization of the Metric Prompt of the Image-caption Pairs.} To better demonstrate the usefulness of the metric prompt in pseudo label filtering, we visualize the metric value of several image-caption pairs. As displayed in Fig. \ref{fig:logits}, we can find that the low-quality image-caption pairs have lower metric prompt values, and the high-quality image-caption pairs have higher metric prompt values. Take the first image-caption pair as an example, the image is about ``a man makes a call in a telephone hall'', but the generated caption is ``a man is shaving a sheep in his mouth'', which is mismatched except for the concept ``a man''. Thus, the metric value has a much lower value ``18.5938''. All these examples fully demonstrate that utilizing metric prompts to filter pseudo labels is reasonable.

%
%
\noindent\textbf{Visualization of the Distance between Semantic Prompts.} The square distances between similar images and different images are illustrated in Fig. \ref{fig:prompt_dis}, respectively. From the figure, we can observe that the square distance between similar images is smaller than the distance between different images. For example, the square distance between two images with a similar ``toilet room'' scene is only 6.67, while the distance is up to 112.74 between the image ``birds flying in the sky'' and the image ``a lot of people with many cupcakes''. The phenomenon of smaller prompt distance existing between images with similar scenes and larger prompt distance existing between images with different scenes demonstrates that the semantic prompt obtained the reasonable semantics of the images.

\noindent\textbf{Visualization of the Generated Captions.} Fig. \ref{fig:captions} shows several representative captions outputted by multiple methods, \textit{i.e.}, VP, VP + SP, PL-UIC(1), and PL-UIC. We can observe that these various UIC models can output reasonable descriptions by using the semantic prompt and the metric prompt. Let us take the first image as an example, the concept ``man'' is described accurately in the VP method, but ``surfing in the ocean on a sunny day'' is imprecise. As for the VP + SP method, the key concept ``slop'' has been correctly recognized. Besides the formerly mentioned concept ``man'', the ``baseball uniform'' and ``bat'' are promisingly described in PL-UIC(1) and PL-UIC, respectively. Especially, the action ``swinging'' is captured accurately by the method PL-UIC. These qualitative samples strongly verify the advantages of the proposed semantic prompt, the metric prompt, and the extraordinary ability of the iteration strategy in caption generator refining.

\section{Conclusion}


We have presented a novel Prompt-based Learning scheme for Unpaired Image Captioning (PL-UIC). By introducing the vision-language pre-trained model (\textit{i.e.}, CLIP), the proposed PL-UIC has for the first time leveraged vision-language prior knowledge in the unpaired image captioning task. To take the advantage of the vision-language alignment in the pre-trained model, two types of prompts are designed, \textit{i.e.}, semantic prompt and metric prompt. The semantic prompt was devised to guide the caption generation for unpaired image captioning in an unpaired supervision fashion. The boosted experimental performance demonstrated that the vision-aware language prior knowledge is effective for generating captions in the UIC task. To further explore the vision-language prior knowledge, the metric prompt was designed to filter pseudo image-caption pairs for the UIC generator refinement in a paired supervision fashion, so the performance of PL-UIC was greatly enhanced.
Extensive experiments demonstrated that the guidance of vision-language prior knowledge is extremely helpful to the UIC task.
It also indicates that it is worthy of focusing on the research of paired image-text transformation for UIC. Overall, we hope that our work will shed light on the development of more effective UIC models.

\bibliographystyle{elsarticle-num}
\bibliography{sample}

\end{document}